%% file: main.tex
\documentclass[letterpaper, 10 pt, conference]{ieeeconf}
\IEEEoverridecommandlockouts
\usepackage{soul}
\usepackage{url}
\usepackage{booktabs}
\usepackage{calc}
\usepackage{graphicx}
\usepackage{multirow}

\usepackage{comment}
\usepackage{paralist}
\usepackage{multicol}
\usepackage{graphicx}
\usepackage{color}
\usepackage{paralist}
\usepackage{amsmath}
\usepackage{amssymb}
\DeclareMathOperator*{\argmax}{argmax}

\usepackage[utf8]{inputenc}
\usepackage{xcolor}
\usepackage{algorithm}
\usepackage{algpseudocode}
\usepackage{mathtools}
\usepackage{amsmath}
\usepackage{tabularx}
    \newcolumntype{L}{>{\raggedright\arraybackslash}X}
\usepackage{array, makecell}
\usepackage{array, makecell}
\algnewcommand\algorithmicforeach{\textbf{for each}}
\algdef{S}[FOR]{ForEach}[1]{\algorithmicforeach\ #1\ \algorithmicdo}

\usepackage{bm}
\usepackage{xspace}
\let\labelindent\relax
\usepackage{enumitem}
\usepackage{pifont}
\usepackage[labelformat=simple, font=small, skip=3pt]{subcaption}
\usepackage[font=small, skip=3pt]{caption}
\usepackage[utf8]{inputenc}
\usepackage{csquotes}
\usepackage{graphicx}
\usepackage{soul,color}
\usepackage{lipsum}
\usepackage[noadjust]{cite}
\usepackage{comment}
\begin{document}
\title{\LARGE \bf Distributed Multi-robot Online Sampling with  Budget Constraints} 
\author{Azin Shamshirgaran$^{1}$ \qquad Sandeep Manjanna$^{2}$\qquad Stefano Carpin$^{1}$%
\thanks{
$^{1}$A. Shamshirgaran and S. Carpin  are with the Department of Computer Science
and Engineering, University of California, Merced, CA, USA.
A. Shamshirgaran is supported by USDA-NIFA under award \# 2021-67022-33452 (National Robotics Initiative).
S. Carpin is partially supported by the IoT4Ag Engineering Research Center funded by the National Science Foundation (NSF) under NSF Cooperative Agreement Number EEC-1941529. Any opinions, findings, conclusions, or recommendations expressed in this publication are those of the author(s) and do not necessarily reflect the view of the U.S. Department of Agriculture or the National Science Foundation.}
\thanks{$^{2}$S. Manjanna is at Plaksha University, Mohali, India.}
}

\IEEEoverridecommandlockouts

\maketitle
\begin{abstract} 
In multi-robot informative path planning the problem is to find a route for each robot in a team to visit a set of locations that can provide the most useful data to reconstruct an unknown scalar field. In the budgeted version, each robot is subject to a travel budget limiting the distance it can travel. Our interest in this problem is motivated by applications in precision agriculture, where robots are used to collect measurements to estimate  domain-relevant scalar parameters such as 
soil moisture or nitrates concentrations.
In this paper, we propose an online, distributed multi-robot sampling algorithm based on Monte Carlo Tree Search (MCTS)  where each robot 
iteratively selects the next sampling location through communication with other robots and considering its remaining budget.

We evaluate our proposed method for varying team sizes and in different environments, 
and we compare our solution with four different baseline methods.
Our experiments show that our solution outperforms the baselines when the budget is tight by collecting measurements leading to smaller reconstruction errors. 
\end{abstract}
\input{sections/introduction.tex}

\input{sections/relatedwork.tex}
\input{sections/background.tex}

\input{sections/method.tex}

\input{sections/results.tex}
\input{sections/conclusion.tex}

\bibliographystyle{plain}
\bibliography{ref.bib}
\end{document}

%% file: sections/introduction.tex
\section{Introduction} 
\label{intro}
\begin{figure}[htb]
      \centering
      \includegraphics[scale=0.07]{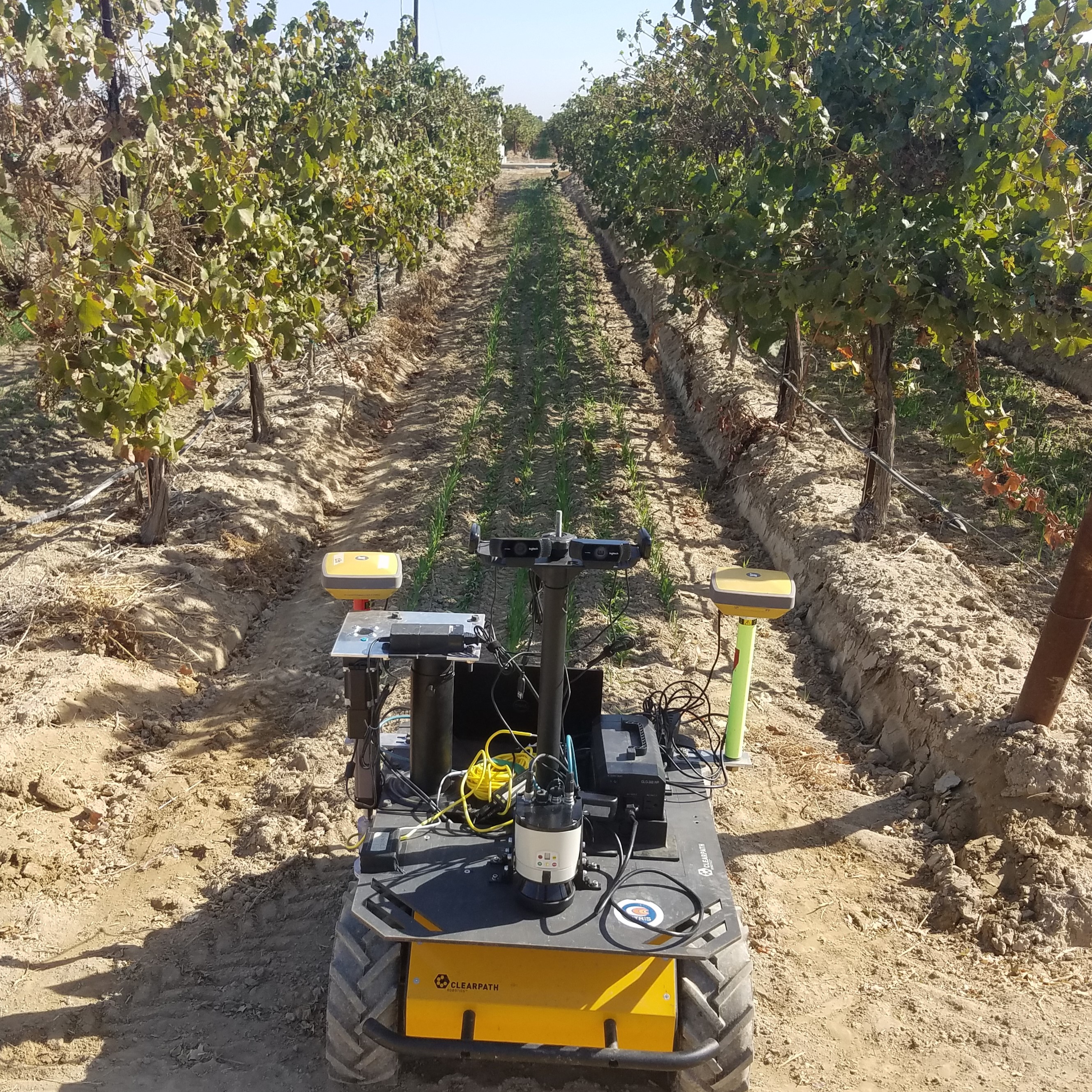}
      \caption{Our robot has been retrofit with a 
      soil moisture sensor that can be inserted in the soil  to collect data
      at preassigned locations, or at locations decided by the robot
      on-the-fly (the soil moisture probe is visible on the left).}
      \label{fig:robot}
   \end{figure}

There is an increasing number of applications for autonomous robots in agriculture~\cite{Stavros2019,Gealy2016},
and  while the most obvious interest may be in fruit harvesting \cite{campbell2022integrated}, there is also sustained demand for 
robots supporting data collection at scale, especially for measuring
and/or estimating scalar parameters such as soil moisture and nitrate concentration
that cannot be easily determined through remote sensing with
satellites and drones (see Fig.~\ref{fig:robot}). 

Robots are expected to play a vital
role in the implementation of farm monitoring systems in support
of \emph{precision agriculture}.  Precision agriculture is defined as
``the matching of agronomic inputs and practices to localized conditions within a field and the improvement of the accuracy of their application'' \cite{FINCH2014}.
Key to this vision is the ability to perform scalable data collection
on demand.
In this context, the capability of deploying a coordinated team of robots to collect data is instrumental, as a team of robots can collect more data per time unit, and also offers increased robustness to individual failures.

For this approach to be  efficient, it is necessary for robots to coordinate their efforts  to avoid unnecessary duplicate work or negative interferences.
As part of the measurement collection task, sample locations can be provided in advance chosen by experts, or randomly, or selected along the way in response to collected data.
Central to our work is the necessity to perform task allocation being aware of the distance a robot can travel before its battery is depleted. From a practical standpoint, a robot running out of energy in the middle of its data collection mission is a major problem, as it  will be necessary to manually recover
it. Our task allocation strategy, 
therefore, focuses both on avoiding duplicate work and on managing  energy constraints.

Communication among robots is another key dimension to be considered in multi-robot scenarios~\cite{bertsekas2021rollout} where in some works all robots can share limited amounts of data with
one another irrespective of the distance~\cite{boothdistributed},  whereas in other approaches
more data is exchanged, but only when robots are sufficiently close to each other ~\cite{manjanna2021scalable}.

 In this work we present
  a distributed algorithm based on communication between robots to avoid having 
 multiple robots collecting measurements at
 the same locations,
  because according to our
 model multiple measurements at the same site do not increase the quality
 of the estimate. Additionally, the proposed algorithm manages exploration and exploitation for each agent by considering the amount  of remaining energy
to prevent the robots from running out of power before a preassigned final recharging location
is reached.  The main contributions of this paper are 1) a budget aware, online distributed method
for multi-robot coordination based on Monte Carlo Tree Search (MCTS); 
2) a strategy to update sampling locations on the fly, to leverage
date acquired during the mission;
3) a thorough comparison with other methods on different benchmark datasets  to evaluate the quality of the proposed method.

The remainder of this paper is organized as follows.
Selected related work is presented in Section \ref{relwork}.
The problem formulation is given in Section \ref{background}, while
our proposed method is presented in Section \ref{proposmeth}.
Extensive simulation results are discussed in Section \ref{res},
and conclusions are offered in Section \ref{conclusions}.

%% file: sections/relatedwork.tex
\section{Related Work} 
\label{relwork}
In this section, we provide pointers to selected works relevant to the 
problem we present in this manuscript.
In multi-robot sampling and monitoring, sectioning and Voronoi partitioning are common approaches. 
In~\cite{kemna2017multi}, the authors propose a dynamic Voronoi approach, where robots repeatedly compute Voronoi partitions and each robot performs sampling within its partition. Although this method shares the exploration task efficiently between robots, the iterative recomputation of Voronoi regions may
lead to many unnecessary motions that are problematic in our application where
robots are subject to a limited travel budget.

In~\cite{thayer2018multi}, we studied multirobot routing in a vineyard 
using a formulation based on the team orienteering problem. In the orienteering problem 
(OP) an agent is required to traverse a graph in which each vertex has a predetermined reward and each edge has a fixed cost. A path should be computed to maximize the sum of collected rewards while ensuring that the sum of the costs of traversed edges does not exceed a preassigned budget. Team Orienteering Problem (TOP) is a version of OP in which a team of robots works together to collect rewards. The main limitation of this line of research is that it assumes that rewards are predetermined in advance. However, in many situations this is not realistic, as data collected on the fly may lead to revised estimates about the value of reaching a certain location.

Recently, MCTS has gained a great deal of attention for multi-robot informative
path planning. In~\cite{jang2021fully, best2019dec}, the authors proposed
a method combining Gaussian processes (GP) and MCTS to solve the problem of environmental monitoring. In~\cite{manjanna2021scalable, manjanna2018reinforcement},  a decentralized approach is proposed using a policy gradient method for multirobot environmental monitoring and sampling. To encourage robots to spread away from other robots, each
robot has a reward function dependent on its distance from the others. 
Furthermore, the method considers a communication range between robots to exchange locations and the history of previous locations with co-working robots. In~\cite{pan2022marlas}, a distributed adaptive sampling method for multi-agent scenarios was proposed which is robust to the failure of robots and communication. To estimate the optimal policy,  deep neural networks and policy gradient methods are used. These works, however, do not explicitly incorporate limits on the distance traveled by robots.

In~\cite{shamshirgaran2022reconstructing}, we presented an offline path planning method based on Q-learning to solve the sampling problem for a single robot in a stochastic environment subject to a preassigned constraint on the distance it can travel.
In~\cite{boothdistributed}, we instead considered the problem of reconstructing a spatial field using multiple robots, Gaussian processes, and MCTS. As part of this work, robots communicate with one another and send their current location and observations to other robots
and explore the spatial properties of  GPs to attempt to spread the robots in
the environment.
In this paper,  we extend this line of research by adding  the ability 
to select new sampling locations as well as considering 
 other robots' actions during planning.



%% file: sections/background.tex
\section{Background and Problem Formulation}
\label{background}

\subsection{Informative Path Planning} 
 We start defining the multi-robot informative path planning (IPP) 
 problem we study in the following. 
 Let $\mathcal{U} \subset \Re^2$ be the environment of interest. 
By visiting a set of locations and collecting samples, our goal is to estimate a scalar function
 $h: \mathcal{U} \rightarrow \mathbb{R}$ which represents
 a parameter of interest (e.g., soil moisture).
 We assume
 there are $\kappa$ robots in the team, each indicated as $R_i \mbox{ with }  i \in [1,2,3,...,\kappa]$.
The  location of each robot ($x,y$-position) is defined by $s_{s}^{R_i}=(s_{x}^{R_i},s_{y}^{R_i})$.
All robots start from  pre-assigned  start positions  $s_{init}^{R_i}$
(e.g., the points where the robots are deployed), and
must terminate  their mission at assigned final locations $s_f^{R_i}$ (e.g., where 
their batteries will either be 
swapped or recharged).  
Each robot has a predetermined travel budget $B^{R_i}$
limiting the distance it can travel. This constraint models, for example, the limited
energy provided by the battery onboard the robot.
There are 
$n$  sample locations of interest\footnote{In  agricultural applications these
preassigned locations are often identified a-priori by domain experts based on past experience and/or data such as yield, plant stress, and the like.} in  $\mathcal{U}$ denoted by 
the set $\mathcal{V} = \{s_1,s_2,\dots,s_n\}$.
The number of locations and their placement in $\mathcal{U}$ is such
that no robot has a sufficient budget to visit all of them.
The goal of multi-robot informative path planning (IPP) is to 
select a path for each robot ${R_i}$ to
visit a subset of locations in  $\mathcal{V}$ 
such that none of the robots exceeds the travel budget and the 
quality of the collected information is maximized. Although $\mathcal{V}$ is given in advance, 
to allow for greater flexibility the set $\mathcal{V}$  can also be updated on the fly, i.e., new sample locations
can be determined by the robots as the mission unfolds.
When a robot reaches a location in $\mathcal{V}$ it uses
its onboard sensor to measure the parameter of interest, 
and this value is then 
used to estimate the unknown function $h$.
More formally, assuming $\rho^{R_i}$ is a path taken by robot ${R_i}$ 
 visiting a subset of sample locations of $\mathcal{V}$, $f(\rho^{R_i})$ is a generic function measuring the quality of
the estimate of $h$ obtained from measurements collected at the
locations visited along the path $\rho^{R_i}$.
Let $C(\rho^{R_i})$ be the travel cost associated with 
traversing $\rho^{R_i}$ for robot ${R_i}$. The multi-robot informative path planning problem (IPP)
can then be expressed as the problem of solving the following
constrained optimization problem for each robot

\[ 
\rho_{\ast}^{R_i} = \argmax_{\rho^{R_i} \in \psi^{R_i}} f(\rho^{R_i}) \, \, \textrm{s.t.} \, C(\rho^{R_i}) \le B^{R_i}
\]
 where $\psi^{R_i}$ is the set of all paths from $s_{init}^{R_i}$ (start location) to $s_f^{R_i}$ (final location) of robot ${R_i}$. 
 In our problem setting, for given path $\rho^{R_i}$, the cost 
 $C(\rho^{R_i})$ is not deterministic, but rather a random variable
 whose realization is obtained only at run time. This models the fact that
 the energy or time spent to move between two locations is in general
stochastic, and we assume that the robot has a probability 
density function describing such  random variable. Hence, while robot $R_i$
moves along $\rho^{R_i}$ from location to location, it is
necessary to monitor 
the energy spent to ensure it does not exceed $B^{R_i}.$
 
\subsection{Gaussian Process Regression}
We model the spatial distribution of the scalar field $h$ being
estimated using Gaussian Processes (GP).
GPs are extensively used in geostatistics \cite{stein1999interpolation, suryan2020learning} to model
environmental parameters (in geostatistics this approach is known 
as kriging.)
We denote
with $x_g^{R_i}$ the scalar reading collected by robot ${R_i}$ when sampling 
location is $s_g \in \mathcal{V}$, and will denote with $\chi_g^{R_i}$
the random variable modeling $x_g^{R_i}$ that follows a Gaussian distribution with  mean $\mu_g$ and variance $\sigma_g^2$.
Using the data collected at the sample locations, 
a posterior of $h$ can be estimated
using standard GP regression algorithms (the reader is
referred to ~\cite{rasmussen2003gaussian} for a comprehensive
introduction to this topic, including GP regression algorithms.)


  \subsection{Monte Carlo tree search (MCTS)}
   \label{mctsall}
 MCTS is an online method for solving sequential,
 stochastic decision making problems. 
  MCTS  builds a tree with a root node representing the current state and edges connecting states that can be reached by executing a single action.  Nodes subsequently added to the tree
 represent states that can be reached through a sequence of actions
 originating at the root. Each action is assigned a $Q$-value
 representing \emph{how good} the action is, which is
 an estimate of the value that will be obtained through
 a complete execution starting with that action.
Once the tree has been constructed, an action is selected from those available at the root. Upon execution of the selected action, the tree is discarded and rebuilt with the next state as its root. A basic version of MCTS consists of the following four steps~\cite{sutton2018reinforcement, carpin2022solving}:
\begin{itemize}
    \item \textbf{Selection:} Using the so-called \emph{tree policy}, a path from the root to a  leaf node is selected.
    \item \textbf{Expansion:} From the selected leaf node, one or more child nodes are added to the tree.
    \item \textbf{Rollout}: A complete episode is simulated from the selected leaf node, or from one of its newly added child nodes (if any). During this simulation, a simple, suboptimal 
    policy is used to decide the actions.
    \item \textbf{Backup:} Based on the return generated by the simulated episode, the action values attached to the tree edges traversed by the tree policy are updated, or initialized.
\end{itemize}

A critical component  is the tree policy for
action selection (``selection'' step in the list above).
 One popular criterion for action selection is  the $UCB$ rule defined in Eq.~\eqref{eq15} and first introduced ~\cite{kocsis2006bandit} 
 Each candidate action $a$ is assigned a $UCB(a)$ value defined as 
 
 \begin{equation}
    \label{eq15}
    UCB(a) = Q_t(a) + c \sqrt{  \frac{\ln t}{N_t(a)}}
\end{equation}
and eventually the action with the highest $UCB$ value is selected
for execution. In Eq.~\eqref{eq15},
$Q_t(a)$ denotes the action value estimate, $N_t(a)$ is the number of times that action $a$ has been selected prior to time $t$, and $c$ is a constant controlling the exploration. Initially, $N_t(a)$ is zero for all actions\footnote{$UCB(a)$ is assumed to be $\infty$ when $N_t(a)=0$,
thus forcing exploration.}. Every time $a$ is selected, $t$ and $N_t(a)$ increase, and every time $a$ is not selected, $t$ increases but not $N_t(a)$, ensuring that all actions will eventually be selected, but actions with lower value estimates or those that have already been selected frequently will be selected less frequently. This criterion
 balances exploration and exploitation, with the balance  determined by
the parameter $c.$

%% file: sections/method.tex
 \section{Proposed Algorithm}
 \label{proposmeth}

Starting from the problem formulation,
we here describe our proposed method.
For each robot $R_i$,
let $\mathcal{A}^{R_i} \subseteq \mathcal{V}$ be the set containing the locations visited by robot $R_i$. 
Initially  $\mathcal{A}^{R_i} = \{s_{init}^{R_i}\}$  and, by 
definition of $\mathcal{A}^{R_i},$ throughout the execution of the
algorithm the current location of the robot $s_{s}^{R_i}$ is one of the elements of $\mathcal{A}^{R_i}$. Each set  $\mathcal{A}^{R_i}$ 
is iteratively expanded by the execution of an action $a$ representing a possible next sampling location  $s_g$ in $\mathcal{V}$ for robot ${R_i}$. The execution of
$a$ implies that the robot will move to $s_g$
and collect a sample at that location.
Using MCTS, the goal is 
to select a \textit{good} sequence of sample locations, $\mathcal{A}^{R_i}$, for robot ${R_i}$ while considering the travel budget, $B^{R_i}$,  
and other robots' decisions.  The meaning of \textit{good} depends on the choice of objective function(s) 
and will be discussed shortly.

  In our proposed method each  robot $R_i$ shares its    visited locations with  other robots 
  with the objective of 
  avoiding having multiple robots visiting the same locations. 
This will lead to
  more distinct samples being collected and ultimately to 
  a better estimate for $h$.
  Our communication model assumes that robots can exchange limited
  information (such as locations) at long range. This is 
   in line with the current technology (e.g., LoRa~\cite{sundaram2019survey})  used by robots in agricultural applications and previous works \cite{boothdistributed,manjanna2021scalable}.

   To select the next location visit,  each robot uses
   a reward function defined as follows.
   For each unvisited location $s_g$, robot $R_i$ defines a
   value $r_{g}^{R_i}$. A possible candidate could be 
  $ r_{g}^{R_i}= \sigma_g^2$ where $\sigma_g^2$ is the variance
   of the posterior estimate of $h$ provided by GP regression
   based on the samples collected up to that moment.
   With this choice, high reward values would be assigned to locations with high uncertainty~\cite{boothdistributed}. In our case,  we scale the predicted variance by the
   distance 
   between the robot's current location and $s_g$
   location. The reward $r_{g}^{R_i}$ associated with a potential sampling location  $s_g$ considered by
   robot $R_i$ is therefore defined as

\begin{equation}
    \label{eq16}
    r_{g}^{R_i} = \frac{\sigma^2_g}{d(s_g,s_s^{R_i})}
\end{equation}
where $\sigma_g$ is variance of the candidate location $s_g \in \mathcal{V}$,   and $d(s_g,s_s^{R_i})$ is the distance between the current location of robot $R_i$ and $s_g$. By introducing the distance into the reward function 
we bias the algorithm to favor closer locations if the predicted variance of  two candidates is the same. The reason for this preference is that we have a limited budget.
The selection of the next location  is performed online, i.e., 
 the reward associated with each location is not predetermined, 
 but re-estimated iteratively based on the locations already visited and data previously collected.
The reward is then used to calculate the expected return, or the action value estimate $Q_t(a)$. In our work we deal with
an \emph{episodic} task, i.e., the task always ends after
a finite amount of time, either because the robot reaches 
the final location, or because it runs out of energy.
In this case, it is typical to define $Q_t(s_g)$ as a function of reward sequence
\begin{equation}
 Q_t(s_g) = r_{g}^{t} + \lambda r_{g'}^{t+1} + ... + \lambda ^{T-t} r_{g''}^{T}; \, \, 0 \le \lambda \le 1 
    \label{eq4}
\end{equation}
where $\lambda$ is a factor discounting future rewards and $T$ is the time of the last action. $g, g', ..., g''$ are the selected sample locations and $r_g$ is the reward associated with the sampling location $g$.
 Algorithm~\ref{algor2} shows the planning algorithm
 independently executed by each robot.

\begin{algorithm}[htb]
    \caption{Online  MCTS based planner for robot $R_i$ with resampling}\label{algor2}
    \begin{algorithmic}[1]
        \State \textbf{Input}: $\mathcal{V}$, $s_{init}^{R_i}$, $s_f^{R_i}$,  $B^{R_i}$
        \State $\mathcal{A}^{R_i} \leftarrow \{ s_{init}^{R_i} \}$
        \State $s_s^{R_i} \leftarrow s_{init}^{R_i}$
        \While{$B^{R_i} > 0$ {\bf and } $s_s^{R_i} \neq s_f^{R_i}$}

             \State $cand^{R_i} \leftarrow \Psi[s_s^{R_i}] \setminus \mathcal{A}^{R_i}$ 
             \State $cand^{R_i} \leftarrow cand^{R_i} \setminus  \mathcal{A}^{R_j}$ for all $j \neq i$
             
             \State $\mathcal{T}, s_g \leftarrow$ MCTS($s_s^{R_i}$, $cand^{R_i}$)
             \State Move to $s_g$, collect reading $x_g^{R_i}$,
             and measure consumed energy $c_s^g$
             
             \State $\sigma_g^2 \leftarrow$ update GP with new observation $x_g^{R_i}$
             \State $\mathcal{V} \leftarrow$ Resampling based on new $\sigma_g^2$
             \State $B^{R_i}\leftarrow B^{R_i}- c_s^g$ 
             
             \State $\mathcal{A}^{R_i} \leftarrow \mathcal{A}^{R_i} \cup \{ s_g \}$
             \State $s_s^{R_i} \leftarrow s_g$
     
     \State Brodacast($s_s^{R_i}$)
     \EndWhile
       \State {\bf return} $ B^{R_i}$, $\mathcal{A}^{R_i}$ 
    \end{algorithmic}
\end{algorithm}
 
The algorithm takes as input the set of  $\mathcal{V}$, the  initial and final locations $s_{init}^{R_i}$
and $s_f^{R_i}$, and the budget $B^{R_i}$ for the each robot.

A children map $\Psi[s_g]$ contains the locations that can be reached from each  location $s_g$. As we consider problem instances with tens or hundreds of possible locations,
considering all of them would lead to search trees with extremely high 
branching factors. Therefore, to
minimize planning time, we limit the set of locations that are considered
from each location, and this set is returned by the function $\Psi$.
More specifically, we limit the branching 
factor to $M$ (an even number). For each location $s_g$, $M/2$ elements in $\Psi$  are the nearest elements in $\mathcal{V}$ and $M/2-1$ are  
randomly chosen from the remaining locations. Moreover the final location, $s_f^{R_i}$ is always added to $\Psi$. This selection balances global exploration and local exploitation. The addition of $s_f^{R_i}$ to $\Psi$
ensures that from any location robot $R_i$ can always consider moving
to the final goal location. This is useful  when the travel budget is about to expire.
$cand^{R_i}$ is the set  of candidate locations for the current location 
and is obtained  by removing already visited locations $\mathcal{A}^{R_i}$ from
the set of reachable locations returned by $\Psi$.

The current location of the robot, $s_s^{R_i}$,  is considered as a root node of the MCTS tree (line 7 in Alg.~\ref{algor2}). 
The MCTS is expanded for a fixed number of iterations. Each time, the path and leaf are chosen using UCB, as per Eq.~\eqref{eq15}. When a leaf is reached, a rollout is executed. During rollout, the planner continues to select
additional random locations until it either reaches the final destination or runs out of energy.  During the MCTS expansion and rollout, every time a candidate location is included in the tree, a generative model is used to estimate how much energy would be consumed.  
This estimate is given by the formula 
\begin{equation}
\label{cost}
  c_s^g = \alpha d(s_{s}^{R_i} , s_{g}) + U(\Lambda)
\end{equation}
where $d(s_{s}^{R_i} ,s_{g})$ is the  distance between the current location, $s_{s}^{R_i}$ and candidate location, $ s_{g}$, and
$U(\Lambda)$ is a random sample from a uniform distribution over
the interval $[0,\Lambda]$.

After the tree $\mathcal{T}$ has been built and the next location, $s_g^{R_i}$, is selected, the budget for robot $R_i$ is updated. Based on the new sample reading, the GP is updated, and based on the updated GP, it will generate a new set of random sample locations with the normalized variance as probabilities associated with each location. In the event that one of the robots reaches its final destination or runs out of energy, other robots will continue their sampling task.

%% file: sections/results.tex
\section{Results and discussion}
\label{res}
\textbf{Methods:} To assess the effectiveness of the proposed method (dubbed RMCTS
in the following), we compared it with four different alternatives.
The first is a non-coordinated MCTS method (NCMCTS), the second is MCTS without resampling (MCST), 
the third is the Orienteering method (Or) \cite{CarpinThayerICRA2018} and the fourth is  Multirobot Planning for Informed Spatial Sampling (MRS)~\cite{manjanna2021scalable}.
NCMCTS  is the same as our proposed method but without any data
shared between robots. MCTS is also the same as our proposed method, but without the resampling step (line 10 in the algorithm). 
The Or method builds a graph with all sample locations and determines the path that collects the maximum reward without exceeding the preassigned budget. In this case,  
 it is necessary to assign a value to each element of $\mathcal{V}$ in advance, consistent with the fact that
 in orienteering one must know the rewards of the vertices beforehand. To make the method comparable with ours, we assigned 
 equal rewards to all vertices,  as our method does not require prior knowledge of the value of collecting a sample at a given location.
 This way, Or will attempt to visit as many locations as possible.
MRS is a decentralized sampling approach where each robot in a team performs an informed survey using a policy-gradient-based sampling strategy. This method uses policy gradient search to directly optimize the policy parameters $\theta$
based on simulated experiences. In its original formulation, this method 
does not consider budget constraints during training. To account for it,
during testing the next location proposed by the algorithm is rejected if there is not enough budget left to reach the designated location and then the robot  moves to the final location $s_f^{R_i}$. In this
 case, the algorithm selects  $s_f^{R_i}$ and tries to reach the final location. The MRS algorithm and its implementation are described in greater detail in~\cite{manjanna2021scalable}.

\textbf{Performance metrics:} To assess the performance of the
various algorithms, we consider two metrics. 
The first is the mean square error between $\hat{h}$ (the estimate of $h$)
and $h$ itself.
In our implementation, GP regression is computed using 
the \emph{scikit-learn} Python library and  its GP regression module   using Matt\'ern Kernel with length scale of 1 and smoothness parameter of $\nu= 1.5$. 
The choice of the kernel and of the parameters was made after having experimentally evaluated different alternatives
and having assessed that these are the best choices.
The second metric is the remaining budget, which is the amount of budget that has not been used when the mission terminates.

\subsection{Synthetic  data-set} 
\label{synthetic}
We start evaluating our proposed method on a bidimensional grid of size $30\times30$
where the scalar field $h$ is defined as a mixture of Gaussian distributions.
The set $\mathcal{V}$ consists of 100 locations. Each robot starts from (0, 0) (upper left corner), and the final location is located at (30, 30) (lower right corner).
We consider different numbers of robots (3 and 5) and different budgets
(100 and 200). For each case, displayed data are averages over 100
independent runs. In all simulations, the MCTS, NCMCTS, and RMCTS algorithms
add 1000 nodes to the tree. For the function $\Psi$ we set $M=30$. In Eq.~\eqref{eq15} we set  $c=3$, in Eq.~\eqref{eq4} we set  $\lambda=1$, and  in   
Eq.~\eqref{cost} we set $\alpha=0.5$ and $\Lambda=1$ and we use Manhattan distance.
In RMCTS, the resampling process is applied every other iteration and after each resampling, $|\mathcal{V}|=30$. %
The MRS method seeks to collect samples at locations with high values for the function $h$ at earlier stages of the exploration. 
During training, MRS runs 20 simulated trajectories to update and learn the parameter of policy $\pi$, and in the test phase it uses that learned policy to generate an explicit action plan.

Table~\ref{table1} summarizes the metrics for MCTS, RMCTS, NCMCTS, and MRS (summary of 100 runs). 
The table displays the initial budget $B$, the  number of robots in a team $N_{R_i}$, the average  MSE error and average remaining budget
for each robot $B_{re}$.
The numerical comparison confirms the superiority of RMCTS across the board.
As the budget and number of robots increase, the performance
of NCMCTS becomes similar to MCTS. Indeed, with a greater number of robots and a large budget, it is possible to visit numerous locations even without coordination. However, this is not the
case when the number of robots is smaller or the budget is tight, and in such cases coordination is key. As a result of resampling, RMCTS achieves a lower MSE because based on the updated GP it generates a better set of sample locations to explore. 
MCTS and RMCTS do not need any prior knowledge about the prior model of the environment whereas MRS needs to know the prior model as a reward map. Also, MRS requires pre-training, while MCTS and RMCTS are online and can choose the next locations on the fly.
To give an order of magnitude, 
training time for one robot in MRS is about 17 minutes, while RMCTS and MCTS do not need training and planning time 
for both is below 15 seconds (cumulative time for all
planning stages alternating with execution).

\begin{figure*}[h!]
\begin{multicols}{4}
      \includegraphics[width=0.8\linewidth]{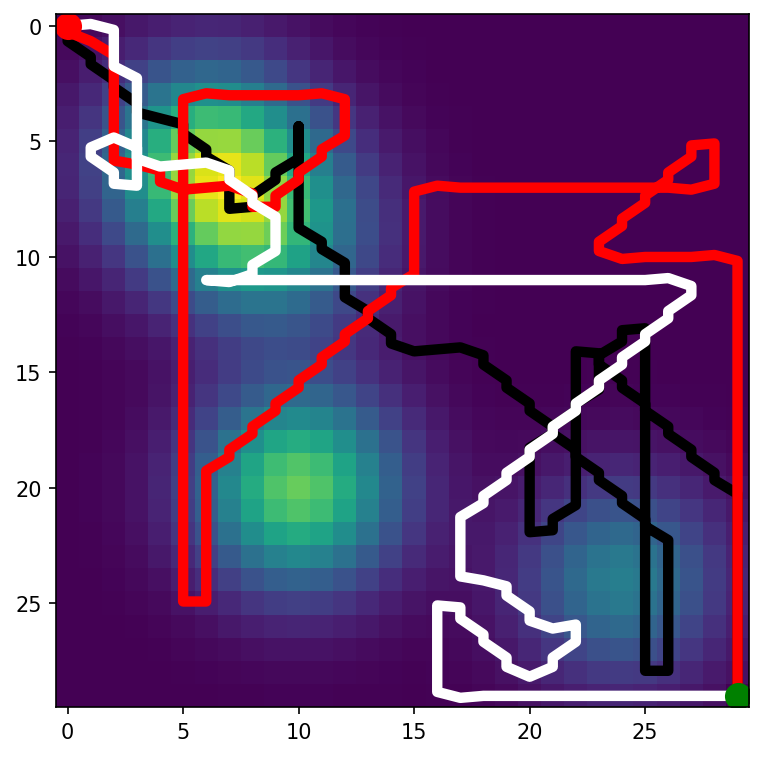}\par \caption*{(a) RMCTS}
        \includegraphics[width=0.8\linewidth]{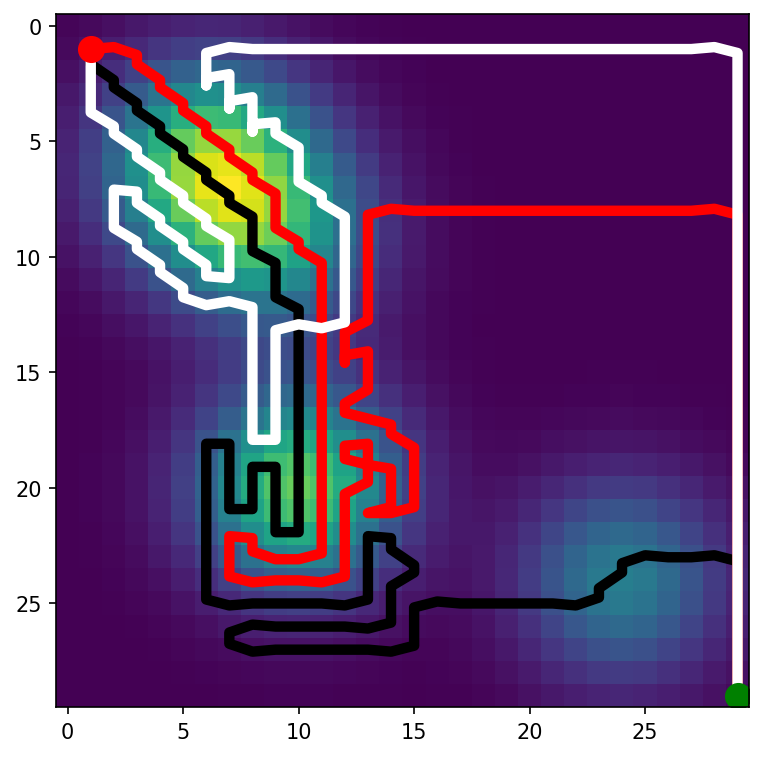}\par \caption*{(b) MRS}
    \includegraphics[width=0.8\linewidth]{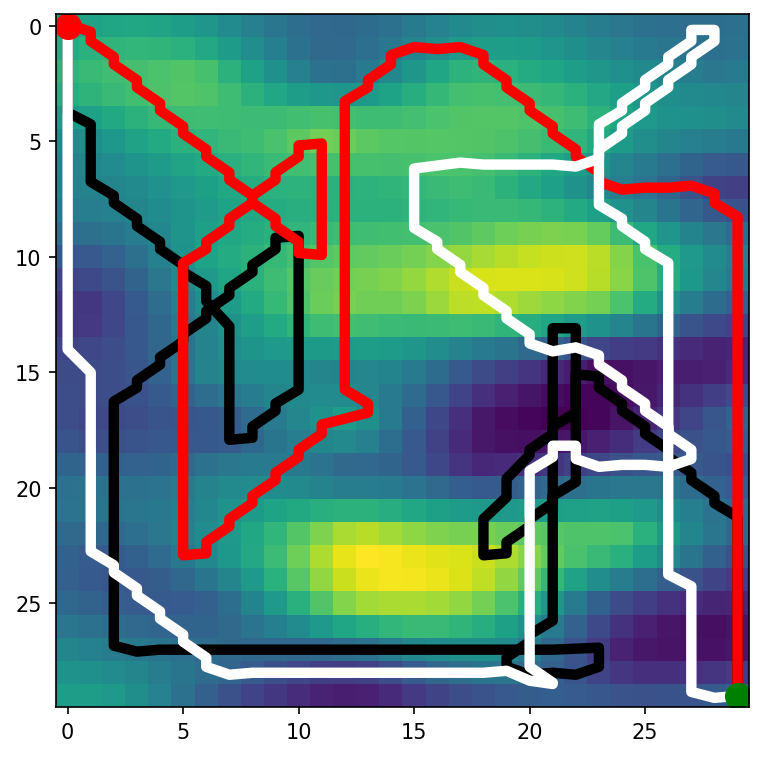}\par \caption*{(c) RMCTS}
    \includegraphics[width=0.8\linewidth]{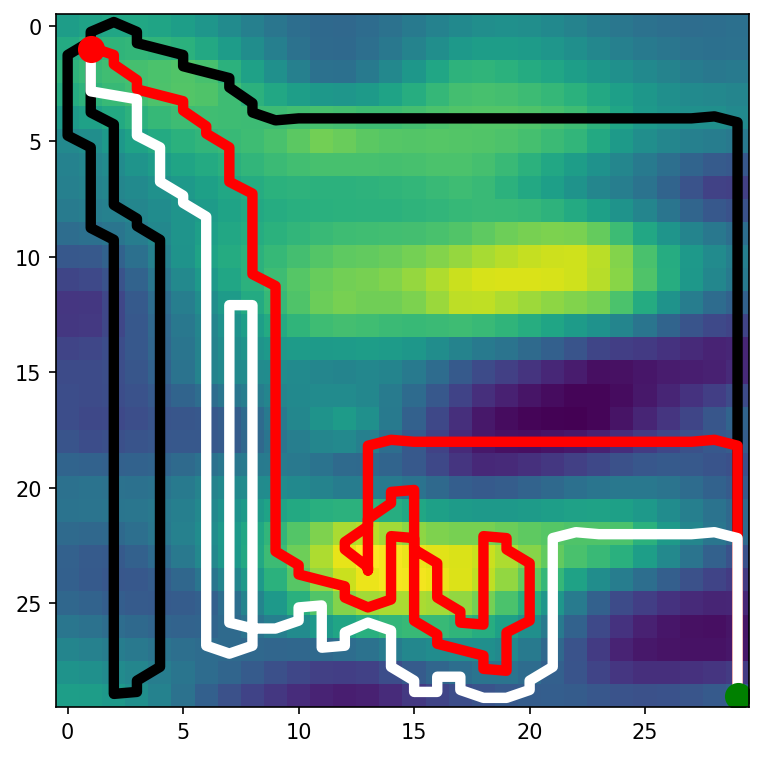}\par \caption*{(d) MRS}

    \end{multicols} 
\begin{multicols}{4}
    \includegraphics[width=0.8\linewidth]{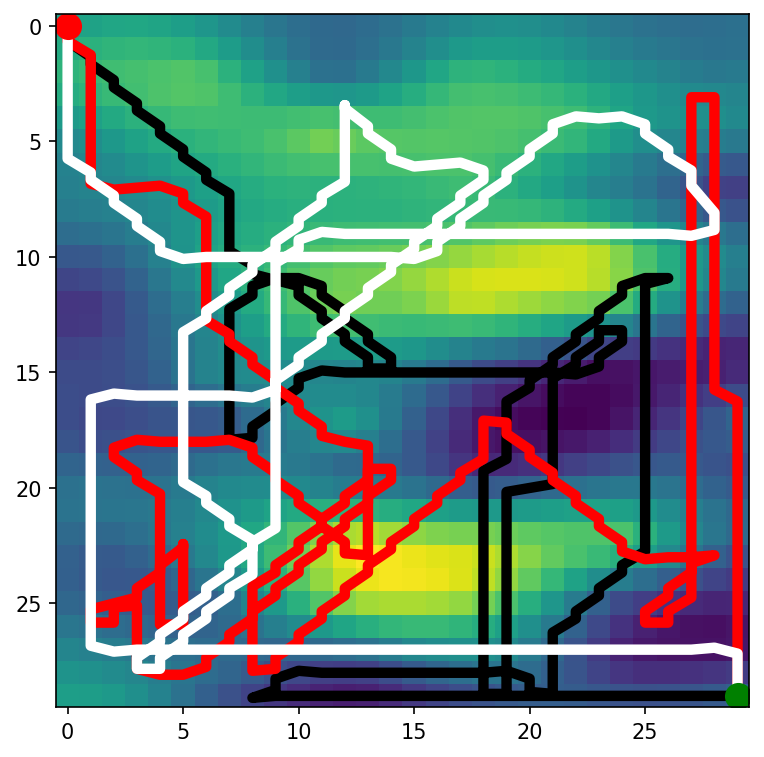}\par\caption*{(e) RMCTS}
        \includegraphics[width=0.8\linewidth]{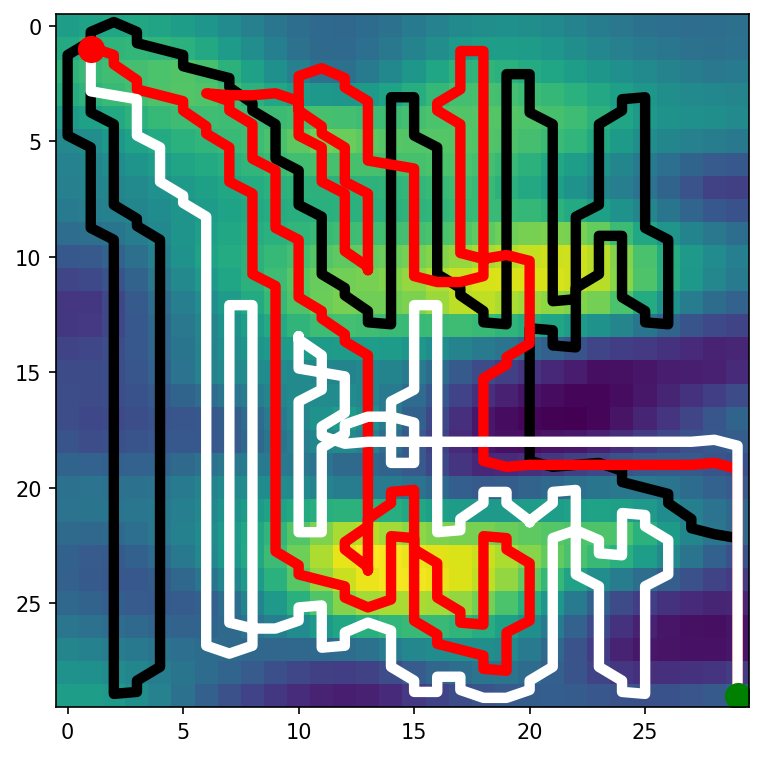}\par\caption*{(f) MRS}
            \includegraphics[width=0.8\linewidth]{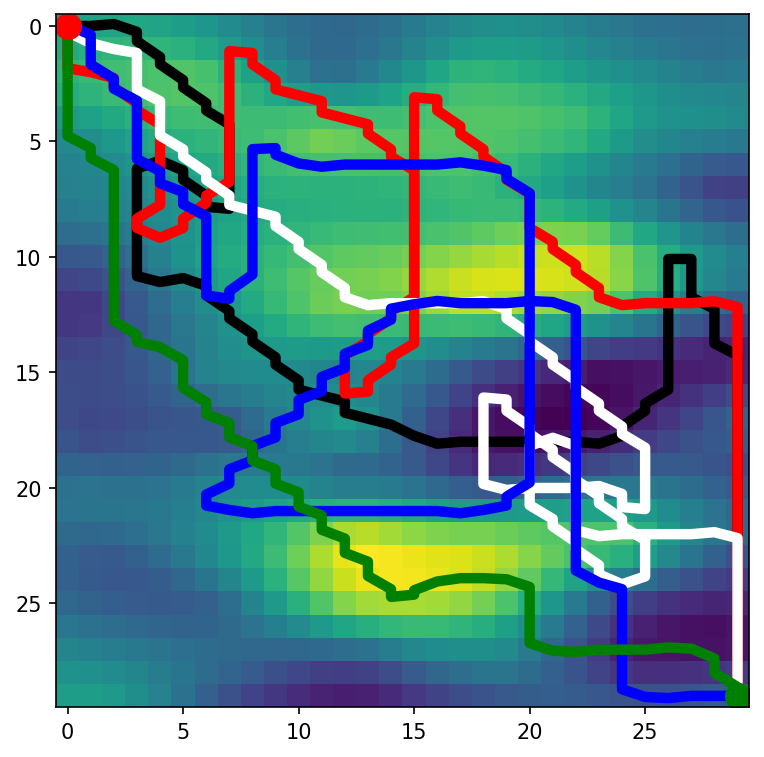}\par\caption*{(g) RMCTS}
    \includegraphics[width=0.8\linewidth]{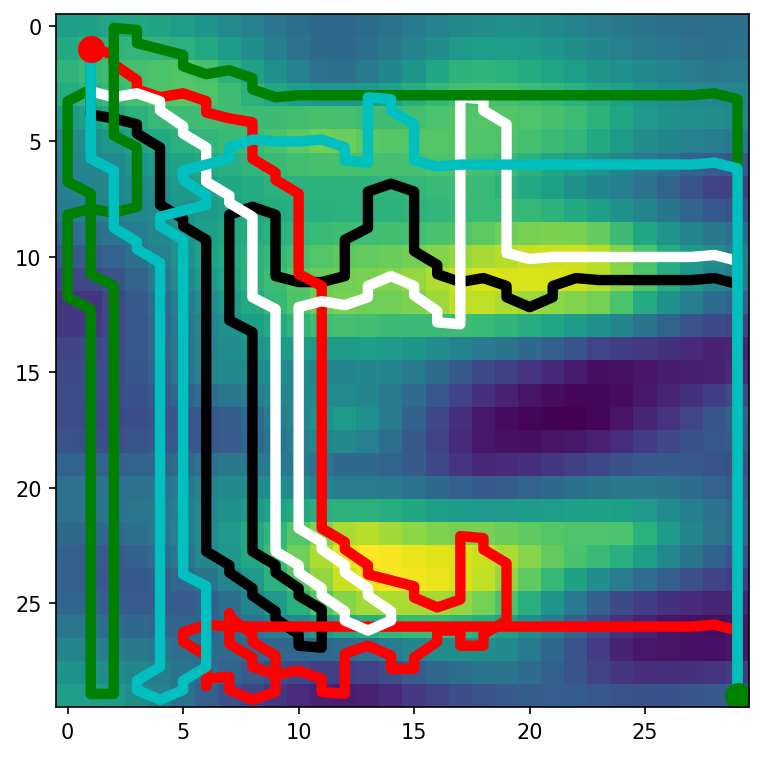}\par\caption*{(h) MRS}
\end{multicols}
\vspace{-4mm}
\caption{ Figures (a)-(b) show three-robots sampling paths with budget $B=100$ in synthetic environment using RMCTS and MRS.
Figures (c)-(d) show three-robots sampling paths with budget $B=100$ in vineyard environment using RMCTS and MRS. Figures (e)-(f) show three-robots sampling paths with budget $B=200$ in vineyard environment using RMCTS and MRS.
 Figures (g)-(h) show five-robots sampling paths with budget $B=100$ in vineyard environment using RMCTS and MRS.}
\label{fig:scenario1}
\end{figure*}

Fig.~\ref{fig:scenario1}(a) and ~\ref{fig:scenario1}(b) show sample paths for 3 robots with $B=100$. 
With MRS, robots focus on areas with high values for the underlying function being reconstructed (warmer colors), while in RMCTS, robots visit all areas. As the goal is to reconstruct the underlying function $h$ with low RMSE error, to do that robots must visit both areas with high and low values.

\begin{table}[hbt!]

\begin{center}\renewcommand\cellalign{lc}
\setcellgapes{3pt}\makegapedcells
\begin{tabular}{ |c|c|c|c|c| } 

 \hline
\hline
$B$& $N_{R_i}$& method&MSE  &$B_{re}$  \\
\hline


100&3&MCTS&0.52&13.72\\ 
100&3&RMCTS&\textbf{0.47}&\textbf{11.23}\\
100&3&NCMCTS&1.27&14.1\\
100&3&MRS &1.39&12 \\

\hline
\hline

100&5&MCTS&0.38&13.21\\
100&5&RMCTS&\textbf{0.35}&11.29\\
100&5&NCMCTS&0.81&15.04\\
100&5&MRS &1.19&\textbf{6}\\

\hline
\hline

200&3&MCTS&0.48&23.72\\ 
200&3&RMCTS&\textbf{0.41}&\textbf{23.13}\\ 
200&3&NCMCTS&0.54&27.11\\
200&3&MRS&1.13&34 \\

\hline
\hline

200&5&MCTS&0.35&21.32\\ 
200&5&RMCTS&\textbf{0.32}&\textbf{18.71}\\ 
200&5&NCMCTS&0.49&20.41\\
200&5&MRS &0.65& 20\\

\hline
\hline

\end{tabular}
\end{center}
\caption{Avg. Results for 100 runs for the synthetic data-set. }
\label{table1}
\end{table}

\subsection{Experimental data-set}
Next, we test all  methods
using a data-set for soil moisture experimentally collected
in a commercial vineyard located in the California Central Valley.
 In this case, we use 100 sample locations that are distributed throughout the environment. The parameters are the same as Sec.~\ref{synthetic}.
 Table~\ref{table2} summarizes the results for 100 runs of MCTS, RMCTS, Or, and MRS methods for one robot. It can be seen that RMCTS outperforms other methods with a tight budget, while MRS achieves better MSE with a higher budget. The Or method consumes almost all of the budget, but it should be noted that for example in our implementation, the planning time for Or (40.21) is more than five times greater than RMCTS (7.52 s) and MCTS (6.18 s). The fact that the MSE is not significantly better in the Or method even with a higher budget is due to the fact that in orienteering, one aims
at collecting the maximum additive reward, and this can
be achieved  by visiting many nearby locations that
will lead to limited additional information to better
estimate the scalar field $h$.

Finally,  table~\ref{table3} summarizes results for 100 runs of MCTS, RMCTS, and MRS methods for teams of three, five, and ten robots. Similar to other cases, we also find that RMCTS outperforms other methods when the budget is tight, while MRS outperforms other methods when the budget is higher. MRS performs better with higher budgets since it considers the samples' locations at each step rather than our method, which considers samples at specific locations. Figure~\ref{fig:scenario1} (c) and (d) show the sampling path for 3 robots with $B=100$ while figure (e) and (f) show the sampling path for 3 robots with $B=200$ and figure (g) and (h) show the sampling path for 5 robots with $B=100$. With RMCTS the robots visit the most informative locations which leads to a more accurate reconstruction of the spatial domain and lower MSE.

\begin{table}[hbt!]

\begin{center}\renewcommand\cellalign{lc}
\setcellgapes{3pt}\makegapedcells
\begin{tabular}{ |c|c|c|c| } 

 \hline
\hline
$B$& method&MSE  &$B_{re}$  \\
\hline


100&MCTS&3.16&18.45\\ 
100&RMCTS&\textbf{2.99}&15.06\\ 
100&Or &4.3&\textbf{5.78} \\
100&MRS &6.67&16 \\
\hline
\hline
200&MCTS&2.99&16.83\\ 
200&RMCTS&2.17&12.03\\ 
200&Or &2.14&\textbf{10.17} \\
200&MRS &\textbf{1.42}&14 \\

\hline
\hline

\end{tabular}
\end{center}
\caption{Avg. Results for 100 runs for the vineyard data-set. }
\label{table2}
\end{table}
\begin{table}[hbt!]
\begin{center}\renewcommand\cellalign{lc}
\setcellgapes{3pt}\makegapedcells
\begin{tabular}{ |c|c|c|c|c| } 
 \hline
\hline
$B$& $N_{R_i}$& method&MSE  &$B_{re}$  \\
\hline
100&3&MCTS&2.83&12.87\\ 
100&3&RMCTS&\textbf{2.50}&\textbf{10.62}\\ 
100&3&MRS &6.67&17 \\
\hline
\hline
100&5&MCTS&2.53&17.38\\
100&5&RMCTS&\textbf{2.41}&\textbf{15.64}\\ 
100&5&MRS &6.23&28\\
\hline
\hline
200&3&MCTS&2.46&20.66\\ 
200&3&RMCTS&2.37&\textbf{12.27}\\ 
200&3&MRS&\textbf{1.08}&13 \\
\hline
\hline
200&5&MCTS&2.38&16.38\\ 
200&5&RMCTS&2.14&14.37\\ 
200&5&MRS &\textbf{0.10}&\textbf{6} \\
\hline
\hline
100&10&MCTS&2.17&27.35\\ 
100&10&RMCTS&1.84&25.11\\ 
100&10&MRS &\textbf{0.23}& \textbf{12}\\
\hline
\hline
\end{tabular}
\end{center}
\caption{Avg. Results for 100 runs for the vineyard data-set. }
\label{table3}
\end{table}


%% file: sections/conclusion.tex
\vspace{-4.5mm}
\section{Conclusions and Future Work}\label{conclusions}
In this paper, we proposed an online distributed multi-robot sampling algorithm based on the MCTS algorithm which is scalable to the size of the team. To minimize revisiting locations,
robots share their past experiences (visited sampling locations). 
A GP model of the scalar field being estimated is updated every time a sample location is measured and is used in the process of generation of a new set of random sample locations. In comparison to baseline methods, our proposed approach is more accurate and makes better use of the limited budget. Also, our proposed method does not require prior knowledge of the environment distribution.  In the future, we intend to incorporate the estimation of other robots' plans into our method and to test the proposed method on the field.